\documentclass{article}
\usepackage[english]{babel}
\usepackage[utf8]{inputenc}
\usepackage{bm}
\usepackage{amsmath,amsthm,amsfonts}
\usepackage{graphicx}
\usepackage{color}
\usepackage{multirow}
\usepackage{subcaption}
\usepackage{fancyhdr}  
\usepackage{geometry}  
\usepackage{caption}

\graphicspath{ {./figs/} }

\date{}

\newcommand{\mytitle}{Reservoir Static Property Estimation Using Nearest-Neighbor Neural Network}

\title{\mytitle \thanks{This note is based on the author's patent CN112396230B.}}

\author{
Yuhe Wang \thanks{National \& Local Joint Engineering Laboratory for Big Data Analysis and Computing Technology, Beijing 100190, China.}
\thanks{Institute for Scientific Computation, Texas A\&M University, College Station, Texas 77843, USA. Email: {\tt yuhe.wang@tamu.edu}.}
}

\date{\normalsize \today}

\begin{document}

\maketitle

\thispagestyle{empty}

\pagestyle{fancy}  

\begin{abstract}
This note presents an approach for estimating the spatial distribution of static properties in reservoir modeling using a nearest-neighbor neural network. The method leverages the strengths of neural networks in approximating complex, non-linear functions, particularly for tasks involving spatial interpolation. It incorporates a nearest-neighbor algorithm to capture local spatial relationships between data points and introduces randomization to quantify the uncertainty inherent in the interpolation process. This approach addresses the limitations of traditional geostatistical methods, such as Inverse Distance Weighting (IDW) and Kriging, which often fail to model the complex non-linear dependencies in reservoir data. By integrating spatial proximity and uncertainty quantification, the proposed method can improve the accuracy of static property predictions like porosity and permeability.  
\end{abstract}

\section{Background}

Reservoir modeling is a critical process in the development of subsurface reservoirs, such as those found in oil and gas fields \cite{ringrose2016reservoir,mi2017enhanced}. Its primary objective is to characterize the spatial distribution of key reservoir properties, including porosity and permeability, which are essential for assessing reservoir reserves, evaluating properties, and determining overall potential \cite{pyrcz2014geostatistical,cipolla2010reservoir}. By integrating data from core samples, well logs, seismic surveys, and other sources, reservoir model offer a detailed representation of the spatial relationships between the essential reservoir properties. This modeling process is not only fundamental for understanding the current condition of the reservoir but also serves as the foundation for subsequent numerical simulations \cite{wang2020generalized, yan2018enhanced} and the development of effective management strategies \cite{cipolla2011seismic,zhang2019novel}.  

Spatial interpolation is a widely used technique in reservoir modeling, involving the estimation of reservoir property distributions across a reservoir based on observations at discrete points \cite{remy2009applied}. In addition to reservoir modeling, spatial interpolation are extensively applied across various domains, including geographic information systems \cite{oliver1990kriging}, hydrology \cite{ly2013different}, and climatology \cite{dobesch2013spatial}. In the context of reservoir modeling, spatial interpolation estimates unknown property values by extrapolating from known data points, such as porosity measurement from well logs and core sample analysis. Numerous spatial interpolation algorithms have been developed, each varying in complexity and accuracy \cite{atkinson2021geostatistical}. 

Inverse Distance Weighting (IDW) is one of the earliest and most straightforward interpolation methods, initially designed from two-dimensional data interpolation \cite{shepard1968two}. Although it has been extended to three-dimensional applications, IDW’s reliance on distance as the sole determinant of spatial relationships limits its capacity to accurately model complex reservoir structures. The method’s simplicity often leads to an oversimplification of spatial dependencies. In contrast, Kriging offers a more advanced approach by incorporating spatial correlations between data points \cite{oliver1990kriging}. As a linear unbiased estimator, Kriging employs a weighted average based on spatial covariance or variograms. However, it typically assumes a static spatial covariance structure across the entire region of interest, which may not adequately capture the inherent variability of geological processes. Additionally, Kriging and other geostatistical techniques often depend on assumptions of linearity and Gaussian distributions, which may not hold true in practical scenarios \cite{remy2009applied, kelkar2002applied}. These limitations restrict the ability of traditional methods to effectively model the complex, non-linear relationships that frequently characterize the spatial distribution of reservoir properties.

In recent years, advancements in machine learning have opened new avenues for tackling the complexities of spatial interpolation. Deep neural networks, in particular, has shown exceptional promise in approximating complex, non-linear functions, making them well-suited for modeling intricate spatial relationships. This potential has been harnessed in a variety of spatial information related tasks, such as groundwater salinity mapping \cite{cui2021gaussian}, regional climate analysis \cite{kadow2020artificial}, well connectivity evaluation \cite{du2020connectivity}, pressure and motion field estimation \cite{du2023novel}, and water saturation distribution prediction \cite{zhang2019potential}. In these applications, neural networks have proven highly effective at capturing the non-linear dependencies that are often inherent in spatial data, thereby offering a more accurate and flexible approach compared to traditional methods. However, in the reservoir modeling context, interpolating reservoir properties presents several challenges, including the limited number of data points typically available in reservoir modeling and the inherently constrained dimensionality of the data \cite{ganguli2024reservoir}. Observation data is often sparse, with measurements obtained at limited number of locations and at resolutions much lower than what is required for accurate reservoir modeling, resulting in significant gaps in spatial coverage. Moreover, the reliance on only a few key features – primarily spatial coordinates and known property values – further limits the overall quality and accuracy of the resulting reservoir models.  

Given the limitation of traditional approaches, there is a pressing need for a better approach that can address the above-mentioned challenges. The method should be capable of handling low-dimensional, sparse datasets while accurately modeling the complex, non-linear spatial relationships within the reservoir. Moreover, it should provide a means to quantify the uncertainty associated with spatial interpolation, thus offering a more reliable framework for reservoir modeling. To this end, this note describes a new kind of method for estimating the spatial distribution of static properties in reservoirs, based on a neighboring neural network approach \cite{wang2019nearest}. This method leverages the strengths of neural networks in modeling non-linear relationships while incorporating spatial proximity data to improve estimation accuracy. Additionally, by quantifying interpolation uncertainty, this approach address the shortcomings of traditional geostatistical methods, offering a more robust solution for reservoir modeling. 

\section{Method}

Method. Given a specific reservoir section where the spatial coordinates and the corresponding static property values, such as porosity and permeability, are available at a limited number of locations, each location can be defined by its spatial coordinates (x, y, z) and associated static property values. In geostatistics, it is well known that fully utilizing information from nearby locations is essential. To this end, a nearest-neighbor algorithm \cite{wang2019nearest} is employed to identify neighboring locations for each location of interest (can be either location with or without known static property values). For example, assuming there are N total locations of interest, the nearest-neighbor algorithm can be used to assign each location m neighboring locations. It is noted that in building the neural network model, the locations with known static property values are used as inputs, whereas during estimation, the goal is to estimate static property values at unknown locations. 

In a basic neural network architecture, random layer is introduced before the existing layers following the input layer to incorporate an element of randomness into the model. The random layer adds variability to the network’s outputs, enabling the generation of multiple realizations of outputs through repeated randomization. By profiling these output realizations, the model can be adapted to quantify the uncertainties associated with the spatial interpolation, providing a more robust and reliable estimation. 

\begin{figure}[h]
    \centering
    \includegraphics[width=0.8\linewidth]{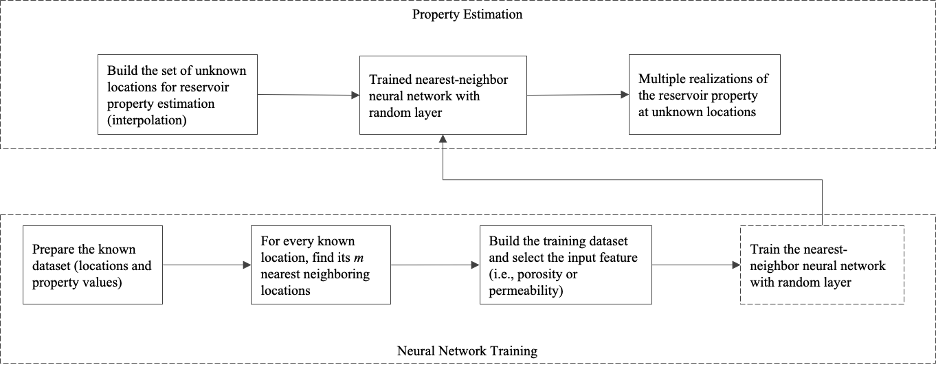}
    \caption{The workflow for estimating reservoir static property using nearest-neighbor neural netwrok}
    \label{fig:workflow}
\end{figure}

Figure \ref{fig:workflow} provides the workflow for reservoir static property estimation using a nearest-neighbor neural network. In the process of training the neural network, the training dataset is composed of the known spatial coordinates of the \(i\)-th location, the known spatial coordinates of the \(m\) neighboring locations of the \(i\)-th location, and the known static property values of these \(m\) neighboring locations, where \(i = 1, 2, 3, \dots, N\). As an example of estimating two-dimensional porosity distribution, the training dataset can be expressed as:

\begin{equation}
\resizebox{0.8\textwidth}{!}{  
$
\begin{bmatrix}
\mathit{X}_1 \\
\mathit{X}_2 \\
\vdots \\
\mathit{X}_N \\
\end{bmatrix}
=
\left[
\begin{array}{ccccccccc}
x_1 & y_1 & x_1^1 & y_1^1 & \Phi_1^1 & \dots & x_1^m & y_1^m & \Phi_1^m \\
x_2 & y_2 & x_2^1 & y_2^1 & \Phi_2^1 & \dots & x_2^m & y_2^m & \Phi_2^m \\
\vdots & \vdots & \vdots & \vdots & \vdots & \ddots & \vdots & \vdots & \vdots \\
x_N & y_N & x_N^1 & y_N^1 & \Phi_N^1 & \dots & x_N^m & y_N^m & \Phi_N^m
\end{array}
\right]
$
}
\end{equation}

$X_1, X_2, \dots, X_N$ represent the samples in the dataset. $(x_i, y_i)$ denotes the spatial coordinates of the $i$-th location, for $i = 1, 2, 3, \dots, N$. The spatial coordinates of the $j$-th neighboring location to the $i$-th location are represented as $(x_{ij}, y_{ij})$, and $\phi_{ij}$ denotes the porosity of the $j$-th neighboring location of the $i$-th location. For the spatial interpolation of different static properties, different input features should be considered. In this example of estimating the spatial distribution of porosity, only the porosity of the location and its $m$ neighboring locations are included as the input feature.

A portion of the dataset is then randomly selected to form the training set, while the remaining samples make up the validation set. Typically, 80-90\% of the samples are used for training, with the remaining reserved for validation. The training samples are used as inputs to the neural network model, and the model’s parameters are updated using the stochastic gradient descent method. After a certain number of iterations of parameter updates, the validation samples are input into the trained model. Then, a ten-fold cross-validation is applied for more comprehensive evaluations.

Once the neural network is trained, the optimal model is used to estimate the distribution of static properties at unknown spatial locations within the selected reservoir section. Let the spatial coordinates of the unknown location be denoted as $(x, y)$. The nearest neighbor algorithm, previously employed, is used to identify the $m$ nearest neighboring locations to the unknown location. The spatial coordinates of the unknown location, along with the spatial coordinates and static property values of its $m$ neighboring locations, are then input into the optimal neural network model to estimate the corresponding static property value for the unknown location. To obtain the full distribution, this estimation process is repeated for each unknown location in the selected reservoir section. Multiple realizations can also be generated for uncertainty quantification through the random layer.

\section{Example}
\begin{figure}[h]
    \centering
    \includegraphics[width=0.6\linewidth]{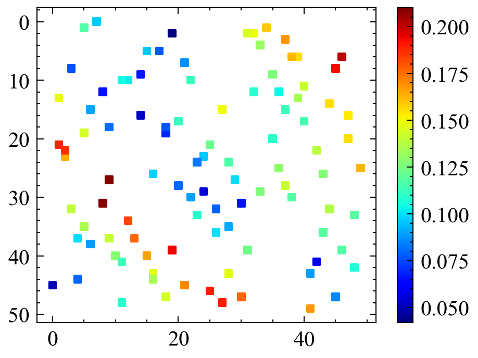}
    \caption{The locations with known porosity values}
    \label{fig:scatter-porosity}
\end{figure}

A simplified square-shaped model is used as an example for illustration to represent a certain layer of a three-dimensional reservoir. The objective is to estimate the porosity ($\phi$) distribution within this layer. In this example, it is assumed that porosity measurements are available at 100 locations, as shown in Figure \ref{fig:scatter-porosity}. Thus, the known dataset for this scenario can be written as:

\[
\left( x_1, y_1, \phi_1 \right), \left( x_2, y_2, \phi_2 \right), \dots, \left( x_{100}, y_{100}, \phi_{100} \right)
\]

Assuming that 15 nearest locations are identified for each known location, then the sample dataset is:

\begin{equation}
\resizebox{0.8\textwidth}{!}{  
$
\begin{bmatrix}
\mathit{X}_1 \\
\mathit{X}_2 \\
\vdots \\
\mathit{X}_{100} \\
\end{bmatrix}
=
\left[
\begin{array}{ccccccccc}
x_1 & y_1 & x_1^1 & y_1^1 & \Phi_1^1 & \dots & x_1^{15} & y_1^{15} & \Phi_1^{15} \\
x_2 & y_2 & x_2^1 & y_2^1 & \Phi_2^1 & \dots & x_2^{15} & y_2^{15} & \Phi_2^{15} \\
\vdots & \vdots & \vdots & \vdots & \vdots & \ddots & \vdots & \vdots & \vdots \\
x_{100} & y_{100} & x_{100}^1 & y_{100}^1 & \Phi_{100}^1 & \dots & x_{100}^{15} & y_{100}^{15} & \Phi_{100}^{15}
\end{array}
\right]
$
}
\end{equation}

\begin{figure}[h]
    \centering
    \includegraphics[width=0.6\linewidth]{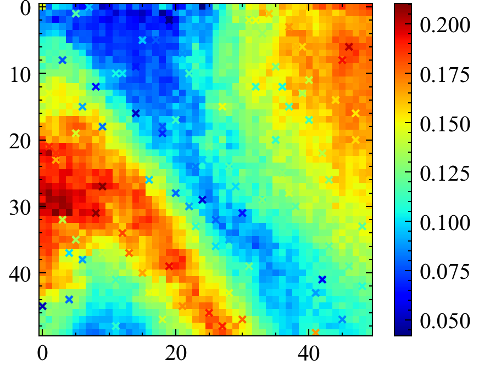}
    \caption{Estimated porosity distribution}
    \label{fig:estimated-porosity}
\end{figure}

Following the steps in Figure \ref{fig:workflow}, with an average validation error toleration of 0.5\%, the obtained porosity distribution is shown in Figure \ref{fig:estimated-porosity}. 

\section{Summary}

. The presented method for estimating reservoir static property distribution uses a nearest-neighbor neural network. This approach effectively captures the non-linear spatial relationships inherent in reservoir geology. As a result, it can improve the quality and robustness of spatial interpolation. The method also introduces multiple randomization processes to quantify the uncertainty in interpolation. This leads to more comprehensive estimation of reservoir static property distributions. However, the current method still only uses spatial coordinates of locations with known property values as input features for the neural network. Incorporating domain-specific knowledge related to spatial interpolation could further enhance its performance, as shown in \cite{zhang2024hybrid}.         

\bibliographystyle{unsrt}
\bibliography{reservoir-static-property-estimation}

\end{document}